\documentclass{article}
\usepackage{spconf,amsmath,graphicx}

\usepackage{enumitem}
\setlist{nosep, leftmargin=14pt}

\usepackage{mwe} %
\usepackage{url, times}
\usepackage{epsfig}
\usepackage{graphicx}
\usepackage{amsmath}
\usepackage{amssymb}
\usepackage[font=small]{subcaption}
\usepackage[dvipsnames]{xcolor}
\newcommand{\venue}[1]{{$_{{\text{#1}}}$}}
\DeclareRobustCommand{\command1}[1]{\textcolor{ForestGreen}{\innervenue{#1}}}
\newcommand{\innervenue}[1]{$_{\text{#1}}$}
\newcommand{\etal}{\textit{et al.}}

 \usepackage{balance}
 \usepackage{multirow}
\usepackage{colortbl}
 \usepackage[colorlinks=true,breaklinks=true]{hyperref}
\usepackage{booktabs}
\usepackage{pifont}

\newcommand{\xmark}{\ding{55}}
\newcommand{\cmark}{\ding{51}}
\newcommand{\bestresult}[1]{\textbf{\textcolor{red}{#1}}}
\newcommand{\secondbest}[1]{\textcolor{blue}{\underline{#1}}}

\title{CodaMal: Contrastive Domain Adaptation for Malaria Detection in Low-Cost Microscopes}
\name{Ishan Rajendrakumar Dave, Tristan de Blegiers, Chen Chen, Mubarak Shah}
\address{Center for Research in Computer Vision, University of Central Florida, Orlando, USA\\
ishandave@ucf.edu, deblegiers.tristan@gmail.com, \{chen.chen, shah\}@crcv.ucf.edu 
}
\begin{document}
\maketitle
\begin{abstract}
  Malaria is a major health issue worldwide, and its diagnosis requires scalable solutions that can work effectively with low-cost microscopes (LCM). Deep learning-based methods have shown success in computer-aided diagnosis from microscopic images. However, these methods need annotated images that show cells affected by malaria parasites and their life stages. Annotating images from LCM significantly increases the burden on medical experts compared to annotating images from high-cost microscopes (HCM). 
  For this reason, a practical solution would be trained on HCM images which should generalize well on LCM images during testing. While earlier methods adopted a multi-stage learning process, they did not offer an end-to-end approach. In this work, we present an end-to-end learning framework, named \textit{CodaMal} (COntrastive Domain Adpation for MALaria). In order to bridge the gap between HCM (training) and LCM (testing), we propose a domain adaptive contrastive loss. It reduces the domain shift by promoting similarity between the representations of HCM and its corresponding LCM image, without imposing an additional annotation burden.
 In addition, the training objective includes object detection objectives with carefully designed augmentations, ensuring the accurate detection of malaria parasites. 
 On the publicly available large-scale \textit{M5-dataset}, our proposed method shows a significant improvement of \textbf{16}\% over the state-of-the-art methods in terms of the mean average precision metric (mAP), provides \textbf{21}$\times$ speed improvement during inference and requires only \textbf{half} of the learnable parameters used in prior methods. Our code is publicly available: \url{https://daveishan.github.io/codamal-webpage/}. 

\end{abstract}
\section{Introduction}
The predominant factor contributing to mortality from malaria is the postponement in its diagnosis and subsequent treatment. Prompt identification is pivotal not just for avoiding health complications but also for mitigating its transmission in communities. Records indicate that malaria primarily impacts individuals in resource-limited tropical and sub-tropical areas where healthcare infrastructures are less robust~\cite{sultani2022towards}.

To enhance the scalability of malaria diagnosis in resource-limited areas, low-cost microscopes (LCM) offer a viable solution. Despite being over 70\% cheaper than high-cost microscopes (HCM), LCMs are constrained by a limited field of view (FOV) and produce less distinct images due to inferior quality lenses~\cite{sultani2022towards}. %
Since LCMs are widely available, an optimal solution should effectively cater to these devices.

\begin{figure}
    \centering
    \includegraphics[width=1\columnwidth]{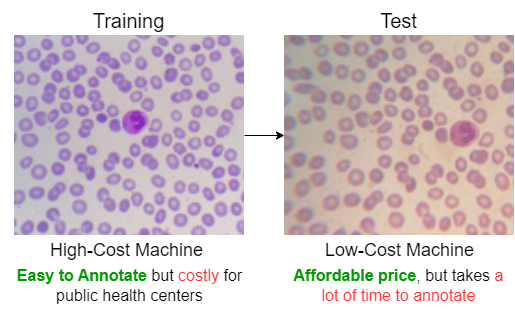}
    \vspace{-5mm}
    \caption{In a practical setup of computer-aided malaria detection: high-cost microscopes (HCM) facilitate ease of annotation for training, whereas low-cost microscopes (LCM) provide an affordable solution for testing in low-resource areas. To address this domain gap, we propose a domain adaptive contrastive loss-based framework, \textit{CodaMal}.}
    \label{fig:motivation}
\end{figure}

Addressing the subjectivity in diagnosis and the scarcity of expert medical personnel, numerous computer-aided microscopic malarial image analysis techniques have emerged. Among these, deep learning-based approaches~\cite{umer2020novel,Hung_2017_CVPR_Workshops, xu2020cross,saito2019strong,chen2018domain} have demonstrated superiority over prior methods that rely on hand-crafted visual features~\cite{dave2017computer, dave2017image, molina2020sequential, fatima2020automatic}. While deep learning techniques excel in interpreting challenging microscopic images, they require annotated data. Medical experts are required to annotate the location and life stage of the malaria parasite within these images. This annotation process becomes extremely time-consuming and burdensome when dealing with LCM images, suggesting that annotating HCM images are more practical due to their clearer view of the cells and life stages of parasites.

Integrating these malaria diagnosis challenges, \textit{a practical deep learning approach should \textbf{train on HCM images} due to easier annotation, and crucially, ensure reliability during \textbf{testing on LCM images} given their prevalence} (see Fig.~\ref{fig:motivation}).

\begin{figure*}[h]
    \centering
    \begin{subfigure}{0.75\linewidth}
    \centering
    \includegraphics[width=\linewidth]{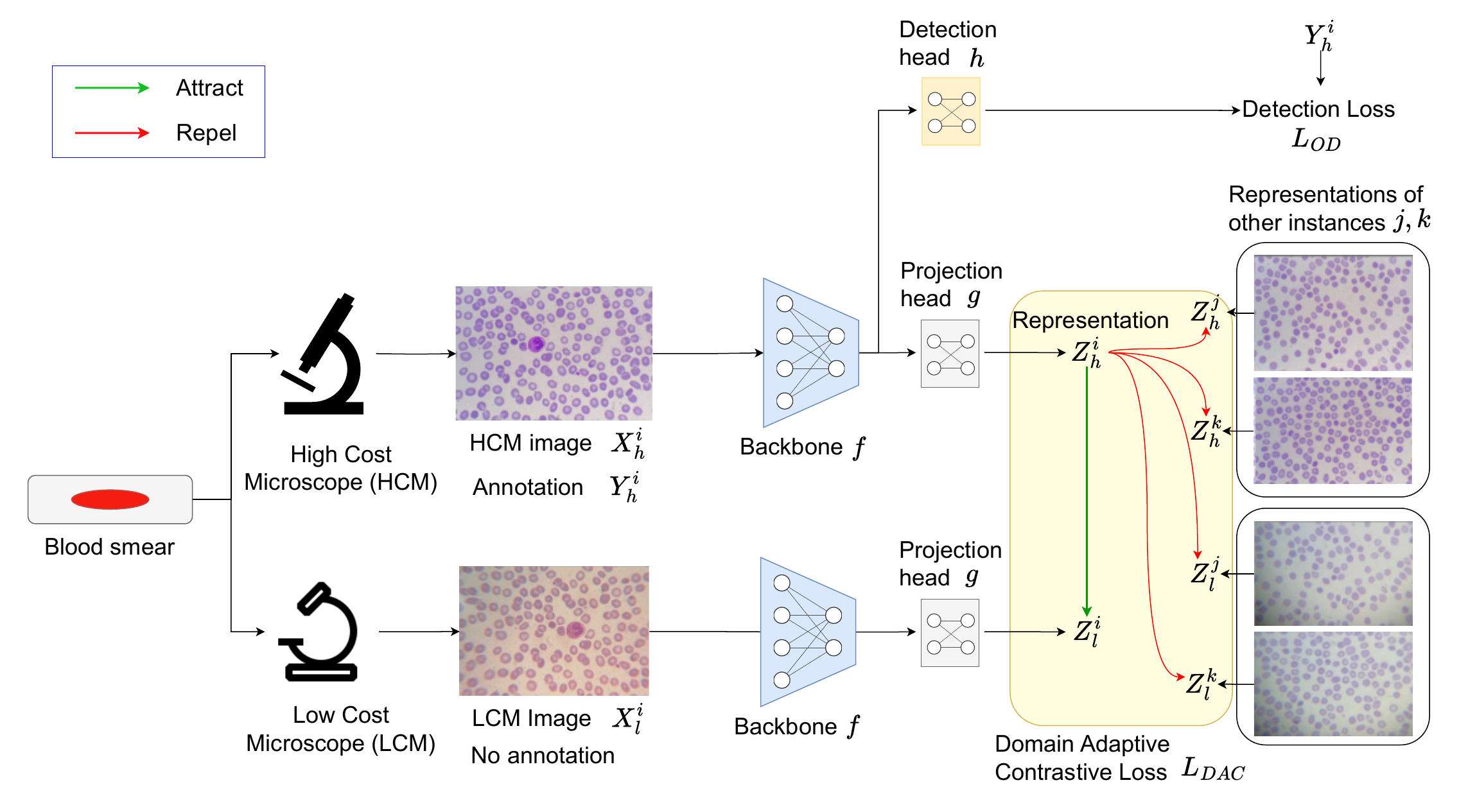}
    \vspace{-2mm}
    \caption{\textbf{Training}}
    \end{subfigure}
    \hfill
    \begin{subfigure}{0.22\linewidth}
    \centering
    \includegraphics[width=0.95\linewidth]{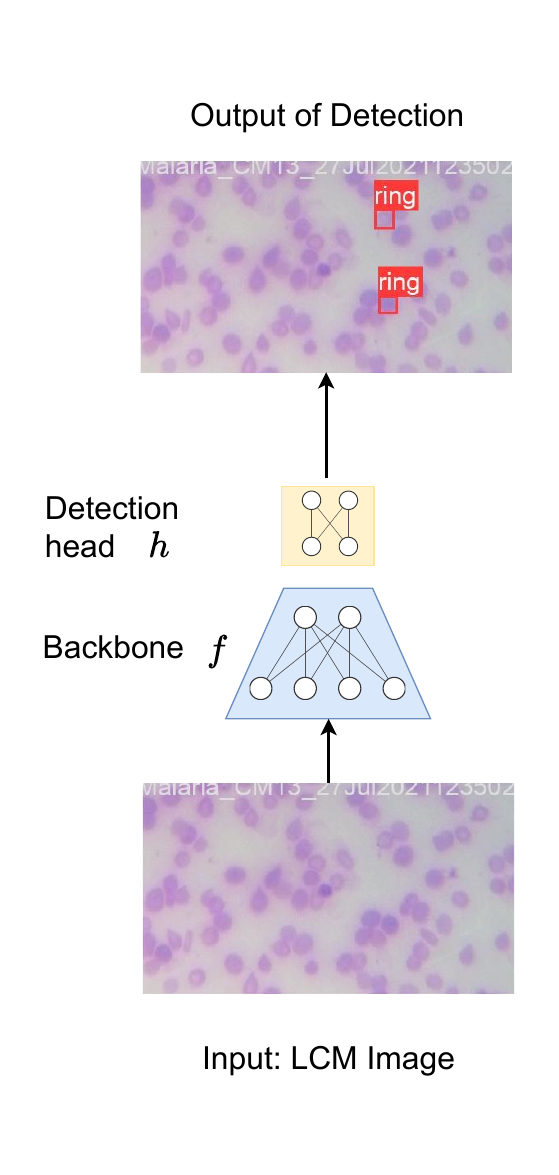}
    \vspace{-2mm}
    \caption{\textbf{Inference (testing)}}
    \end{subfigure}
    \vspace{-0.5mm}
    \caption{Schematic diagram of our end-to-end method for malaria detection. (a) During training, an HCM image ($X^i_h$) and its annotations ($Y^i_h$) are paired with the corresponding unannotated LCM image ($X^i_l$) that is captured from the same field-of-view, instance $i$. The HCM image ($X^i_h$) 
    is utilized to train the backbone ($f$) and its detection head ($h$) using the standard object detection losses ($L_{OD}$). For domain adaptation, both $X^i_h$ and $X^i_l$ are processed through the same backbone ($f$) and non-linear projection head ($g$). The HCM representation, $Z^i_h$, is encouraged to maximize similarity with the corresponding LCM instance ($Z^i_l$) (indicated by the \textcolor{ForestGreen}{green arrow}), while minimizing similarity with the representations of other LCM and HCM image instances ($j$,$k$) i.e. instances captured from different FOV or different blood smear (highlighted by the \textcolor{red}{red arrow}). (b) Once trained, the model receives LCM images and predicts the location and life-stage of the malaria parasite as its final output.
    }
    \label{fig:main}
\end{figure*}

In light of these observations, Sultani~\etal~\cite{sultani2022towards} introduced the M5-dataset, capturing images from both LCM and HCM. It is the largest publicly available dataset for malaria diagnosis. They benchmarked numerous object detection baselines across varied network architectures and training methodologies. Their proposed solution adopts a multi-stage approach: an initial stage which involves training a generative model to generate a pseudo-LCM image from its HCM counterpart followed by a subsequent stage that employs a triplet loss to minimize the domain gap between the generated LCM image and the genuine LCM image. Although their method outperforms the prior object detection baselines, a notable drawback of their strategy is its reliance on multi-stage training, lacking a seamless end-to-end method to address domain shifts, i.e., HCM (training) to LCM (testing).

To address these challenges, in this paper, we propose   \textit{CodaMal} (COntrastive Domain Adpation for MALaria). 
For the domain shift, we propose the Domain Adaptive Contrastive loss ($L_{DAC}$), promoting representational %
similarity between LCM and HCM images. Our proposed framework embraces a single-stage training process, delivering an end-to-end solution that markedly simplifies training complexity and reduces model training time. Adhering to the testing protocol established by ~\cite{sultani2022towards}, we evaluate our method on the M5-dataset. Our approach substantially outperforms the previous state-of-the-art ~\cite{sultani2022towards}, by showing improvements of 16\% on 1000$\times$ magnification FOV and 15\% on the 400$\times$ FOV in terms of mean average precision (mAP). Furthermore, our method does not require additional training data or more trainable parameters compared to the prior methods.

\section{Related Work}
\noindent\textbf{Prior Work in Computer-Aided Diagnosis}
Previous efforts have introduced various microscopic image datasets for malaria, such as BBBC041\cite{BBBC041}, Malaria655\cite{boray2010}, MPIDB\cite{loddo2018mp}, and IML\cite{arshad2021dataset}. However, these datasets do not include blood smear images from microscopes of varying costs (i.e., HCM \& LCM). This limitation makes the M5 dataset\cite{sultani2022towards} the most appropriate for addressing this issue, as it provides images taken with both HCM and LCM. 

\noindent\textbf{Domain Adaptation in Computer Vision}
Contrastive learning methods~\cite{simclr, moco, byol, tclr}, known for learning powerful label-free representations, have been beneficial for domain adaptation~\cite{surveyCODA}. To the best of our knowledge, our research is the first to investigate contrastive domain adaptation between HCM and LCM, for improving malaria parasite detection.
\section{Method}
The training set \( D_{train} \) is constructed by capturing the same field-of-view of a smear using both LCM and HCM. Each HCM image \( X^i_h \) is annotated with the location and life stages of the malaria parasites \( Y^i_h \). We also have the LCM image \( X^i_l \) corresponding to the same view, which could be slightly misaligned due to equipment mechanical precision and lens distortion (More details about acquisition in ~\cite{sultani2022towards}).  In this study, we focus on the four distinct life stages of P. Vivax: ring, trophozoite, schizont, and gametocyte. The objective is to train a deep-learning model \( f (\cdot) \) that minimizes the domain discrepancy between HCM and LCM during training, and subsequently provides accurate detection of malaria parasites on the test set \( D_{test} \), which comprises solely of LCM images. 

\begin{table*}[h]
\centering
\begingroup
\setlength{\tabcolsep}{4.5pt}
\arrayrulecolor[rgb]{0.753,0.753,0.753}
\begin{tabular}{l|c|c|c|cc} 
\arrayrulecolor{black}\toprule
\arrayrulecolor[rgb]{0.753,0.753,0.753}
\multicolumn{1}{c|}{\multirow{2}{*}{\textbf{Method}}} & \multirow{2}{*}{\begin{tabular}[c]{@{}c@{}}\textbf{End-to-End}\\\textbf{Training?}\end{tabular}} & \multirow{2}{*}{\begin{tabular}[c]{@{}c@{}}\textbf{Number of}\\\textbf{Params (M)}\end{tabular}} & \multirow{2}{*}{\begin{tabular}[c]{@{}c@{}}\textbf{Inf. time}\\\textbf{(ms)}\end{tabular}} & \multicolumn{2}{c}{\textbf{HCM $\xrightarrow{}$ LCM}}  \\
\multicolumn{1}{c|}{}                                 &                                                                                                  &                                                                                                  &                                                                                                 & \textbf{1000x} & \textbf{400x}                                      \\ 
\arrayrulecolor{black}\hline
Chen et al.~\cite{chen2018domain}~\venue{CVPR'18}                                       & \cmark                                                                                               & 43.7                                                                                             & 184                                                                                             & 17.6           & 21.5                                               \\
Saito et al.~\cite{saito2019strong}~\venue{CVPR'19}                                      & \cmark                                                                                               & 43.7                                                                                             & 184                                                                                             & 24.8           & 21.4                                               \\
Xu et al.~\cite{xu2020cross}~\venue{CVPR'20}                                         & \cmark                                                                                               & 43.7                                                                                             & 184                                                                                             & 15.5           & 21.6                                               \\
Faster RCNN                                           & \cmark                                                                                               & 43.7                                                                                             & 184                                                                                             & 17.1           & 26.7                                               \\
Faster RCNN + Synthetic LCM Tuning~\cite{sultani2022towards}                    & \textcolor[rgb]{0.7,0.7,0.7}{\xmark}                                                                                              & 43.7                                                                                             & 184                                                                                             & 33.3           & 31.8                                               \\
Faster RCNN + Synthetic LCM Tuning + Ranking Loss~\cite{sultani2022towards}     & \textcolor[rgb]{0.7,0.7,0.7}{\xmark}                                                                                              & 43.7                                                                                             & 184                                                                                             & 35.7           & 32.4                                               \\
Faster RCNN + Synthetic LCM Tuning + Triplet Loss~\cite{sultani2022towards}     & \textcolor[rgb]{0.7,0.7,0.7}{\xmark}                                                                                              & 43.7                                                                                             & 184                                                                                             & 37.2           & 32.2                                               \\
Sultani et al.~\cite{sultani2022towards}~\venue{CVPR'22}                                   & \textcolor[rgb]{0.7,0.7,0.7}{\xmark}                                                                                              & 43.7                                                                                             & 184                                                                                             & \secondbest{37.5}           & \secondbest{33.8}                                               \\
\rowcolor[rgb]{0.784,0.902,0.976}\textbf{CodaMal (Ours)}                               & \textbf{\cmark}                                                                                      & \textbf{21.2}                                                                                    & \textbf{8.9}                                                                                    & \bestresult{53.6}  & \bestresult{48.7}                                      \\
\arrayrulecolor{black}
\bottomrule
\end{tabular}
\endgroup
\caption{\textbf{Comparison with state-of-the-art approaches}. The performance metric is mAP@0.5 IoU. 1000$\times$ and 400$\times$ represent the magnification FOV of LCM images. Highlighted \bestresult{Red} shows the best results and \secondbest{Blue} shows second best results.}
\label{tab:main}
\end{table*}

Our framework, \textit{CodaMal}, depicted in Fig.~\ref{fig:main}, is built upon two main training objectives. The first objective includes object detection losses, detailed in Sec.~\ref{sec:objdet}, ensuring the accurate detection of malaria parasites. The second, discussed in Sec.~\ref{sec:ldac}, introduces the Domain Adaptive Contrastive Loss ($L_{DAC}$) to minimize the domain discrepancy between HCM and LCM images. Together, these objectives ensure robust parasite detection on LCM during testing.

\subsection{Learning Objectives for Parasite Detection}
\label{sec:objdet}
We employ a single-shot object detection technique using the CSP-DarkNet53~\cite{yolov5} backbone to apply the object detection losses. The input HCM image ($X^i_{h}$) undergoes a series of transformation functions before the losses are applied. These transformations encompass useful augmentations tailored for object detection, such as random cropping, scaling, mix-up, mosaic, etc. The overall object detection loss function, \(L_{OD}\), is a combination of three primary components:
\begin{equation}
L_{OD} = L_{cls} + L_{loc} + L_{obj}
\end{equation}

\begin{itemize}
    \item \(L_{cls}\) is the classification loss, gauging the difference between the predicted and actual object (i.e. 4 life-stages of parasites) classes in the HCM images.
    \item \(L_{loc}\) represents the localization loss, capturing the %
    difference between the predicted bounding box coordinates and the true coordinates of detected objects.
    \item \(L_{obj}\) denotes the objectness loss, assessing whether grid cells in the HCM image contain an object or not.
\end{itemize}

For a more in-depth understanding of each loss component, we refer readers to ~\cite{yolo}.

\subsection{Domain Adaptive Contrastive Loss ($L_{DAC}$)}
\label{sec:ldac}
The Domain Adaptive Contrastive (DAC) loss aims to minimize the domain discrepancy between the HCM and LCM images by encouraging representation %
similarity between the two domains. 

\noindent\textbf{Non-linear Projection Layer} Let $F_{h}^i$ and $F_{l}^i$ denote the feature maps extracted from the P5 layer of CSP-DarkNet53 backbone $f(\cdot)$ for the $i^{th}$ unaugmented HCM image ($X^i_{h}$) and the $i^{th}$ unaugmented LCM image ($X^i_{l}$), respectively. We introduce a non-linear projection layer $g(\cdot)$, which consists of a series of fully connected layers followed by ReLU activation functions. This layer maps the high-dimensional feature maps to a lower-dimensional latent space to obtain representations:
\begin{equation}
\begin{aligned}
F_{h}^i &= f(X_{h}^i), \quad F_{l}^i = f(X_{l}^i), \\
Z_{h}^i &= g(F_{h}^i), \quad Z_{l}^i = g(F_{l}^i).
\end{aligned}
\end{equation}

\noindent where, $Z_{h}^i$ and $Z_{l}^i$ are the non-linear projections of the $i^{th}$ HCM and LCM features, respectively.

\noindent\textbf{Computing the DAC Loss} To compute $L_{DAC}$, we measure the similarity between the non-linear projections of the HCM and LCM features using a cosine similarity function, and then apply a contrastive loss building upon~\cite{simclr}. Formally, this can be described as:
\begin{equation}
L_{DAC} = -\frac{1}{N} \sum_{i=1}^{N} \log \frac{\exp\left(\frac{Z_{h}^i \cdot Z_{l}^i}{\|Z_{h}^i\| \|Z_{l}^i\|} / \tau\right)}{\sum_{j \neq i} \exp\left(\frac{Z_{h}^i \cdot Z_{l}^j}{\|Z_{h}^i\| \|Z_{l}^j\|} / \tau\right)}
\end{equation}

Here, \(N\) is the batch size, and \(\tau\) is a temperature hyperparameter. Positive pairs are made up of corresponding HCM and LCM instances, while negative pairs consist of other instances in the batch. By minimizing this loss, the model is trained to produce invariant features across the HCM and LCM domains, enhancing the generalization of the object detector for LCM images.

\begin{figure*}[h]
    \centering

    \includegraphics[width=\linewidth]{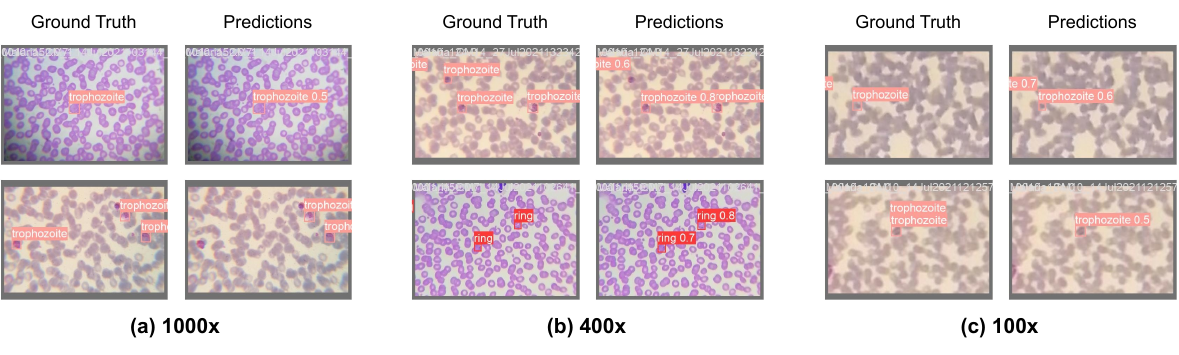}

    \vspace{-3mm}
    \caption{\textbf{Qualitative Results} of our proposed method on LCM test images at 3 magnification scales of the microscope.}
    \label{fig:qualitative}

\end{figure*}
\section{Experiments}

\subsection{Dataset and Performance Metrics}
We use the \textit{multi-microscope multi-magnification malaria (M5) dataset}~\cite{sultani2022towards}, the largest publicly available dataset for this study. It uniquely provides both HCM and LCM images of the same fields, taken at magnifications of 1000$\times$, 400$\times$, and 100$\times$. The M5 dataset contains 7,542 images with 20,331 annotated malarial cells and represents four life stages of P.Vivax: ring, trophozoite, schizont, and gametocyte.

Following ~\cite{sultani2022towards}, we use the same performance metric: mean average precision (mAP) at a 0.5 IoU threshold to compare the overlap between predicted and ground-truth bounding boxes. For context, mAP is calculated by averaging the average precision (AP) for each class, where AP measures the area under the precision-recall curve~\cite{lin2014microsoft:conf_typical}. We also report precision and recall results.

\subsection{Implementation Details}
We utilize 2-layer MLP as following~\cite{simclr} as our $g(\cdot)$ and utilize standard detection head $h(\cdot)$ following ~\cite{yolo}. More details can be found from our publicly available codebase\footnote{\url{https://github.com/DAVEISHAN/CodaMal}}.

\subsection{Results}
Following the standard protocol from ~\cite{sultani2022towards}, we provide results on two FOV magnification settings: (1) training on 1000$\times$ HCM and testing on 1000$\times$ LCM images. (2) training on 400$\times$ HCM and testing on 400$\times$ LCM images.
As shown in Table~\ref{tab:main}, our method, \textit{CodaMal}, significantly outperforms prior best results by \textbf{16\%} and \textbf{15\%} relatively on 1000$\times$ and 400$\times$ of FOV in terms of mAP and sets a new state-of-the-art. Also, it is worth noting that our model is \textbf{21}$\times$ faster during inference and requires only \textbf{half} of the learnable parameters used in prior methods. Inference time in the table is reported using NVIDIA T4 GPU on Google Colab. Since all prior methods utilize a Faster R-CNN backbone, they share the same number of parameters and inference time.

The key contributor of our higher performance is employing an end-to-end optimization for domain adaptation, in contrast to the prior best approach\cite{sultani2022towards}, which relies on a two-stage process. Initially, they generate synthetic LCM images using a GAN model, followed by using these images for domain adaptation. The GAN model, lacking feedback from the domain adaptation stage, produces suboptimal performance. Regarding inference speed, our method benefits significantly from using a single-shot detector, as opposed to ~\cite{sultani2022towards} which depends on a region-proposal network(RPN) that is based on a 2 stage detector.

Besides the mAP results, our method also achieves a precision of 0.8761 and Recall of 0.4571, on 1000$\times$ magnification FOV setting. Qualitative results are shown in Fig.~\ref{fig:qualitative}, with high-resolution visualizations available on our webpage\footnote{\url{https://daveishan.github.io/codamal-webpage/}}.

\subsection{Ablations}

\noindent\textbf{Effect of $L_{DAC}$}: The incorporation of the Domain Adaptive Contrastive (DAC) loss in our method yields notable improvements in performance as shown in Table~\ref{tab:dac}. When applied to the 1000$\times$ FOV images, the DAC loss contributes to a \textbf{7.8\%} increase in accuracy. This enhancement is even more pronounced for the 400$\times$ FOV images, where the DAC loss results in a remarkable \textbf{20.2\%} improvement. These outcomes underline the effectiveness of the DAC loss in bridging the domain gap between high-cost and low-cost microscope images and promoting invariance, leading to better malaria detection across varying magnification levels.
\begin{table}[h]
\arrayrulecolor{black}
\centering
\refstepcounter{table}
\label{table_1}
\begin{tabular}{lll} 
\toprule
                     & \textbf{1000$\times$}          & \textbf{400$\times$}           \\ 
\hline
without $L_{DAC}$ & 49.7                   & 40.5                    \\
with $L_{DAC}$  & \textbf{53.6~\command1{(+7.8\%)}} & \textbf{48.7~\command1{(+20.2\%)}~}  \\
\bottomrule
\end{tabular}
\caption{Ablation for Domain Adaptive Contrastive Loss}
\label{tab:dac}
\end{table}

\noindent\textbf{Effect of different backbone initialization} Following the prior object detection works, we also initialize our backbone $f(\cdot)$ by weights from MS-COCO~\cite{lin2014microsoft:conf_typical} and compare it with the random initialization point as shown in Table~\ref{tab:init}. 
It is worth noting that our method which is initialized with random weights (43.1\%) also outperforms previous state-of-the-art~\cite{sultani2022towards} (37.5\%) which utilizes pretrained weights from \cite{lin2014microsoft:conf_typical}.

\begin{table}[h]
\centering
\begin{tabular}{ll} 
\toprule
\textbf{Backbone Initialization} & \multicolumn{1}{l}{\textbf{mAP}}  \\ 
\hline
Random initialization              & 43.1                              \\
MS-COCO~\cite{lin2014microsoft:conf_typical}                 & \textbf{53.6~\command1{(+24.36\%)}}                             \\
\bottomrule
\end{tabular}
\caption{Ablation for various model initializations}

\label{tab:init}
\end{table}

\section{Conclusion}
In conclusion, our study presents a novel end-to-end trainable method for the more practical setting of deep learning learning-based malaria parasite detection: training on HCM and testing on LCM images. The incorporation of the domain adaptive contrastive loss bridges the domain gap by promoting invariance between HCM and LCM domains. Our method, \textit{CodaMal}, substantially outperforms the prior best methods up to 44\% and sets a new state-of-the-art with 21$\times$ speedup in inference. Our open-source code will facilitate future developments in this field for the community.

These findings not only highlight the potential of our method to enhance the early and accurate diagnosis of malaria but also emphasize the importance of domain adaptation in improving the generalization capabilities of deep learning-based models for microscopy applications. 
Future research directions may explore the application of our method to other microscopy-based diagnostic tasks, further refining the method and assessing its impact on a broader range of diseases and detection scenarios.

\bibliographystyle{IEEEbib}
\bibliography{egbib}

\begin{thebibliography}{10}

\bibitem{sultani2022towards}
Waqas Sultani, Wajahat Nawaz, Syed Javed, Muhammad~Sohail Danish, Asma Saadia, and Mohsen Ali,
\newblock ``Towards low-cost and efficient malaria detection,''
\newblock in {\em 2022 IEEE/CVF Conference on Computer Vision and Pattern Recognition (CVPR)}. IEEE, 2022, pp. 20655--20664.

\bibitem{umer2020novel}
Muhammad Umer, Saima Sadiq, Muhammad Ahmad, Saleem Ullah, Gyu~Sang Choi, and Arif Mehmood,
\newblock ``A novel stacked cnn for malarial parasite detection in thin blood smear images,''
\newblock {\em IEEE Access}, vol. 8, pp. 93782--93792, 2020.

\bibitem{Hung_2017_CVPR_Workshops}
Jane Hung and Anne Carpenter,
\newblock ``Applying faster r-cnn for object detection on malaria images,''
\newblock in {\em Proceedings of the IEEE Conference on Computer Vision and Pattern Recognition (CVPR) Workshops}, July 2017.

\bibitem{xu2020cross}
Minghao Xu, Hang Wang, Bingbing Ni, Qi~Tian, and Wenjun Zhang,
\newblock ``Cross-domain detection via graph-induced prototype alignment,''
\newblock in {\em Proceedings of the IEEE/CVF Conference on Computer Vision and Pattern Recognition}, 2020, pp. 12355--12364.

\bibitem{saito2019strong}
Kuniaki Saito, Yoshitaka Ushiku, Tatsuya Harada, and Kate Saenko,
\newblock ``Strong-weak distribution alignment for adaptive object detection,''
\newblock in {\em Proceedings of the IEEE/CVF Conference on Computer Vision and Pattern Recognition}, 2019, pp. 6956--6965.

\bibitem{chen2018domain}
Yuhua Chen, Wen Li, Christos Sakaridis, Dengxin Dai, and Luc Van~Gool,
\newblock ``Domain adaptive faster r-cnn for object detection in the wild,''
\newblock in {\em Proceedings of the IEEE Conference on Computer Vision and Pattern Recognition}, 2018, pp. 3339--3348.

\bibitem{dave2017computer}
Ishan Dave and Kishor Upla,
\newblock ``Computer aided diagnosis of malaria disease for thin and thick blood smear microscopic images,''
\newblock in {\em 2017 4th International Conference on Signal Processing and Integrated Networks (SPIN)}. IEEE, 2017, pp. 561--565.

\bibitem{dave2017image}
Ishan Dave,
\newblock ``Image analysis for malaria parasite detection from microscopic images of thick blood smear,''
\newblock in {\em 2017 International Conference on Wireless Communications, Signal Processing and Networking (WiSPNET)}. IEEE, 2017, pp. 1303--1307.

\bibitem{molina2020sequential}
Angel Molina, Santiago Alf{\'e}rez, Laura Bold{\'u}, Andrea Acevedo, Jos{\'e} Rodellar, and Anna Merino,
\newblock ``Sequential classification system for recognition of malaria infection using peripheral blood cell images,''
\newblock {\em Journal of Clinical Pathology}, vol. 9, no. 10, pp. 665–670, 2020.

\bibitem{fatima2020automatic}
Tehreem Fatima and Muhammad~Shahid Farid,
\newblock ``Automatic detection of plasmodium parasites from microscopic blood images,''
\newblock {\em Journal of Parasitic Diseases}, vol. 44, no. 1, pp. 69--78, 2020.

\bibitem{BBBC041}
``Bbbc041, malaria dataset,'' \url{https://bbbc.broadinstitute.org/BBBC041}.

\bibitem{boray2010}
F~Boray Tek, Andrew~G Dempster, and Izzet Kale,
\newblock ``Images of thin blood smears with bounding boxes around malaria parasites (malaria-655),''
\newblock .

\bibitem{loddo2018mp}
Andrea Loddo, Cecilia Di~Ruberto, Michel Kocher, and Guy Prod’Hom,
\newblock {\em MP-IDB: The Malaria Parasite Image Database for Image Processing and Analysis}, pp. 57--65,
\newblock 2019.

\bibitem{arshad2021dataset}
Qazi Arshad, Mohsen Ali, Saeed-Ul Hassan, Chen Chen, Ayisha Imran, Ghulam Rasul, and Waqas Sultani,
\newblock ``A dataset and benchmark for malaria life-cycle classification in thin blood smear images,''
\newblock {\em Neural Computing and Applications}, vol. 34, 2022.

\bibitem{simclr}
Ting Chen, Simon Kornblith, Mohammad Norouzi, and Geoffrey Hinton,
\newblock ``A simple framework for contrastive learning of visual representations,''
\newblock in {\em Proceedings of the International Conference on Machine Learning}, 2020.

\bibitem{moco}
Kaiming He, Haoqi Fan, Yuxin Wu, Saining Xie, and Ross Girshick,
\newblock ``Momentum contrast for unsupervised visual representation learning,''
\newblock in {\em Proceedings of the IEEE/CVF Conference on Computer Vision and Pattern Recognition}, 2020, pp. 9729--9738.

\bibitem{byol}
Jean-Bastien Grill, Florian Strub, Florent Altch{\'e}, Corentin Tallec, Pierre Richemond, Elena Buchatskaya, Carl Doersch, Bernardo Avila~Pires, Zhaohan Guo, Mohammad Gheshlaghi~Azar, et~al.,
\newblock ``Bootstrap your own latent-a new approach to self-supervised learning,''
\newblock {\em Advances in Neural Information Processing Systems}, vol. 33, pp. 21271--21284, 2020.

\bibitem{tclr}
Ishan Dave, Rohit Gupta, Mamshad~Nayeem Rizve, and Mubarak Shah,
\newblock ``Tclr: Temporal contrastive learning for video representation,''
\newblock {\em Computer Vision and Image Understanding}, p. 103406, 2022.

\bibitem{surveyCODA}
Poojan Oza, Vishwanath~A. Sindagi, Vibashan~Vishnukumar Sharmini, and Vishal~M. Patel,
\newblock ``Unsupervised domain adaptation of object detectors: A survey,''
\newblock {\em IEEE Transactions on Pattern Analysis and Machine Intelligence}, pp. 1--24, 2023.

\bibitem{yolov5}
Glenn Jocher, Alex Stoken, Jirka Borovec, NanoCode012, Yang Zhang, Ayush Chaurasia, Yonghye Kwon, and Stijn Verdenius,
\newblock ``Yolov5: State-of-the-art object detection,'' \url{https://github.com/ultralytics/yolov5}, 2020.

\bibitem{yolo}
Joseph Redmon, Santosh Divvala, Ross Girshick, and Ali Farhadi,
\newblock ``You only look once: Unified, real-time object detection,''
\newblock in {\em Proceedings of the IEEE Conference on Computer Vision and Pattern Recognition}, 2016, pp. 779--788.

\bibitem{lin2014microsoft:conf_typical}
Tsung-Yi Lin, Michael Maire, Serge Belongie, James Hays, Pietro Perona, Deva Ramanan, Piotr Doll{\'a}r, and C~Lawrence Zitnick,
\newblock ``Microsoft coco: Common objects in context,''
\newblock in {\em European Conference on Computer Vision}. Springer, 2014, pp. 740--755.

\end{thebibliography}

\end{document}